\documentclass[journal]{IEEEtran}

\usepackage{amssymb}
\usepackage{graphicx}
\usepackage{mathptmx}
\usepackage{amsmath}
\usepackage{amssymb}
\usepackage{amsthm}
\usepackage{adjustbox}
\usepackage{caption}
\usepackage{subcaption}
\usepackage{algorithmic,algorithm}
\usepackage{float}
\usepackage{booktabs}

\theoremstyle{definition}

\begin{document}


\title{A Proactive Management Scheme for Data Synopses at the Edge}

\author{\IEEEauthorblockN{Kostas Kolomvatsos\IEEEauthorrefmark{1}, Christos Anagnostopoulos \IEEEauthorrefmark{2}} \\

\IEEEauthorblockA{\IEEEauthorrefmark{1} Department of Informatics and Telecommunications, University of Thessaly, Papasiopoulou 2-4, 35131, Lamia Greece}\\
\IEEEauthorblockA{\IEEEauthorrefmark{2} School of Computing Science, University of Glasgow, Lilybank Gardens 18, G12 8RZ, Glasgow UK}\\

\thanks{Manuscript received July, 2021. 
Corresponding author: K. Kolomvatsos (email: kostasks@uth.gr).}

}

\maketitle

\begin{abstract}
The combination of the infrastructure provided by the Internet of Things (IoT) with numerous processing nodes present at the Edge Computing (EC) ecosystem opens up new pathways to support intelligent applications. Such applications can be provided upon humongous volumes of data
collected by IoT devices being transferred to the edge nodes through the network. 
Various processing activities can be performed on the discussed data
and multiple collaborative opportunities between EC nodes can facilitate the execution of the desired tasks.
In order to support an effective interaction between edge nodes, the knowledge about the geographically distributed data should be shared. 
Obviously, the migration of large amounts of data will harm the stability of the network stability and its performance. 
In this paper, we recommend the exchange of data synopses than real data between EC nodes to provide them with the necessary knowledge about peer nodes owning similar data. 
This knowledge can be valuable when considering decisions such as data/service migration and tasks offloading.
We describe an continuous reasoning model that builds a temporal similarity map of the available datasets to get nodes understanding the evolution of data in their peers. 
We support the proposed decision making mechanism through an intelligent similarity extraction scheme based on an unsupervised machine learning model, and, at the same time, combine it with a statistical measure that represents the trend of the so-called discrepancy quantum.
Our model can reveal the differences in the exchanged synopses and provide a datasets similarity map which becomes the appropriate knowledge base to support the desired processing activities. 
We present the problem under consideration and suggest a solution for that, while, at the same time, we reveal its advantages and disadvantages through a large number of experiments.
\end{abstract}


\IEEEpeerreviewmaketitle

\section{Introduction}
\label{introduction}
The current advent of the Internet of Things (IoT)
and the functioning of numerous devices close to 
end users create many opportunities for novel applications that will facilitate their
activities \cite{najam}. 
This huge infrastructure can easily support intelligent pervasive applications.
IoT devices exhibit  
low computational resources, thus, they can only perform
simple processing activities. 
Their connection with the Edge Computing (EC) infrastructure through the 
network gives them the capability of transferring the collected data in an upwards mode towards the 
Cloud datacenters.
In any case, this connectivity opens the room for 
new processing opportunities at the EC.
This means that another processing infrastructure can be located at the EC where 
numerous nodes can support the provision of analytics and other services to 
users or applications \cite{najam}.
We can easily discern two interconnected ecosystems, i.e., the EC and the IoT 
that can execute complex tasks for resulting knowledge.
Recently, operators begun the deployment of massive resources at the EC 
to evolve it into a large, distributed, and capillary computing environment \cite{carrega}.
EC nodes can offer storage and processing solutions
close to the location where data are originally collected.
Knowledge can be represented in the form of predictive analytics queries, tasks, or explanatory-driven models and statistics. 

Any processing performed at the EC faces multiple challenges for supporting novel pervasive applications.
An interesting pathway is the efficient management of the available services and datasets formulated in a geo-distributed manner while the opportunities for a collaborative approach
when executing tasks is also important.
These aspects will give to nodes the necessary ability to react 
in a very dynamic environment and select the optimal line of actions
for any request.
Furthermore, mobility can result in a change on the demand for services or data
and affect the behaviour of EC nodes.
Some research efforts try to 
deal with the dynamic data demand \cite{karanika} or,
proactively, 
select the appropriate hosts to store the incoming data \cite{kolomvatsosCOMPUTING}.
If we focus on the collaboration between EC nodes in cases where there is the need 
of services or data migration as well as offloading some tasks \cite{chen2}, \cite{dinh}, \cite{huang}, \cite{kolomvatsosFGCS}, we have to 
support their behaviour with models that result the `matching' between 
them at every time instance.
One critical decision is to detect the `similarity' between the distributed datasets 
being formulated upon the reports of IoT devices.
It is significant to know the matching between nodes concerning their datasets
in order to support the aforementioned activities and avoiding spending resources for 
delivering `inappropriate' results.

In this paper, we propose the use of a model that enhances the intelligence
of EC nodes through the proactive monitoring scheme.
Our motivation is related to the need of having EC prepared to select the 
appropriate peers when services/data migration or tasks offloading decisions should be made.
We elaborate on a model that exposes the temporal statistical difference 
between synopses of datasets present at different nodes.
In particular, we monitor the so-called data discrepancy quanta, i.e., the difference between pairs
of synopses in order to reveal the evolution of the similarity between
distributed datasets.
We combine an unsupervised machine learning scheme (i.e., clustering) with a 
statistical measurement of the trend of discrepancy quanta. 
The aim is to create a list of similar peers for each EC node and conclude a temporal similarity map
adopted for local decision making.
The proposed approach can be also combined with any other decision making 
for services/data migration, tasks offloading or synopses definition.
The ultimate goal is that EC nodes can save time when the aforementioned decision making should be 
executed as they proactively conclude the similar peers. 
The salient contributions of this paper are:
\begin{itemize}
	\item we define the concept of the discrepancy quantum os the difference between pairs of synopses extracted upon distributed datasets and the corresponding monitoring mechanism;
	\item we adopt a clustering model upon the discrepancy quanta and deliver strong estimations about the evolution of them in a temporal aspect;
	\item we combine the aforementioned machine learning scheme with a statistical approach utilized to expose the statistical trend of the discrepancy quanta;
	\item we elaborate on a simple, however, fast reasoning mechanism to deliver the 
	final list of similar peers locally;
	\item we provide an extensive experimental evaluation and comparative assessment with baseline models showcasing the strengths of our approach.
\end{itemize}

The paper is organized as follows.
Section \ref{related} reviews the related work \& Section \ref{preliminaries} provides an overview of our problem and the proposed mechanism. Section \ref{monitoring} introduces the machine learning and statistical trend analysis methods for realizing the envisioned approach. Section \ref{evaluation} reports on the experimental evaluation and comparative assessment, while Section \ref{conclusions} concludes the paper with future research agenda.

\section{Related Work}
\label{related}
The administration of humongous volumes of data and their migration in the network
may naturally cause bottlenecks and a degradation in the performance.
Data synopses seem to be a convenient way to 
`inform' a group of nodes for the available geo-distributed datasets keeping the amount 
of the exchanged data at the lowest possible levels.
According to \cite{aggarwal}, a synopsis is defined as a `high' level description of the available data depicting their statistics. 
Synopses may be adopted to 
approximate the appropriate response to various queries and estimate their results \cite{Chakrabarti}.
They should be exchanged when a difference with the previously reported versions is detected \cite{kolomvatsostkde}. 
Various techniques have been proposed in the past for the generation of data synopses, e.g., 
sampling \cite{Lakshmi}, load shedding 
\cite{Tatbul}, sketching \cite{Babcock} and micro cluster based summarization \cite{Aggarwal1}.
Sampling incorporates a probabilistic selection of a subset of the actual data while 
load shedding has the goal of dropping some data when the system identifies a high load to 
avoid bottlenecks.
Sketching randomly projects a subset of features that describe the data 
incorporating mechanisms for the 
vertical sampling of the stream and, finally,
micro clustering targets to the management of the multi-dimensional aspect of any data 
towards to the processing of the 
data evolution over time.
Special attention on the performance of the synopsis generation technique 
should be paid when data depict the evolution of a phenomenon, thus, 
the mechanism adopted to deliver synopses should rely on the spatio-temporal axes of the collected datasets \cite{anagnogrouping}, \cite{anagnotson}, \cite{kolomvatsosfusion}, \cite{kolomvatsosiisa}.
Data streams is the appropriate setup to receive and pre-process 
the collected data depicting the phenomenon under consideration though the adoption 
of sensors or autonomous nodes (e.g., unmanned vehicles) 
\cite{kolomvatsosiot}.
Upon these streams, we can easily support the 
extraction of knowledge, the detection of events or any other
processing that may assist in the delivery of efficient services and
applications.
Data and knowledge can be aggregated to support a uniform view
of the delivered results making the extracted knowledge transferable and usable by any processing component that adopts a unified schema \cite{kolomvatsostopk}.
Specific metadata can be also adopted 
to describe the collected data and the aggregated results
\cite{kolomvatsossemantic}. 

Services and data migration are research subjects that recently attracted the attention of researchers.
The interested reader can find an interesting survey in \cite{wangsurvey}.
Any migration activity has the intention to move services or data at the location where they are required
to support the requestors.
Migration activities enhance the processing capabilities of nodes 
and close the potential `gaps' in the available data.
Another reason behind the significance of migration is the creation of a fault tolerant system.
However, the transfer of huge volumes of services/data in the network may jeopardize its stability.
It is convenient if services/data are migrated 
with different levels of granularity across heterogeneous 
nodes where virtualized resources are present \cite{bellavista}.
This will facilitate, e.g., the migration of parts of a service 
that should be incorporated in the processing capabilities of the local nodes.
Data migration is adopted in various research efforts like \cite{bellavista2}. 
The discussed publication elaborates on a machine learning model and the LibSVM toolkit to estimate the mobility of users and, accordingly,
perform a proactive data migration to meet their needs.
The authors of \cite{alam} propose the use of a reinforcement learning scheme combined with a 
multi-agent environment.
The target is to build on the autonomous nature of agents enhance them with the ability of 
getting decisions that will lead to the optimal reward.
Autonomous entities can also be part of a swarm and through a collaborative approach can conclude the optimal line of actions to deliver 
the best possible outcome \cite{kolomvatsosPSO}.
In \cite{devita}, deep learning is utilized to decide upon data migration.
The goal is to learn the hidden characteristics of the system and 
get the best possible decisions.
Another effort based on machine learning is presented in \cite{alfarraj}.
In this research activity, machine learning assists in 
the aggregation of data when migration is concluded.
Game theory also provides a technique to support migration, e.g., 
the authors of  
\cite{zhang} discuss a coalitional game based pricing scheme to reason on 
the offloading relationship between data and processing nodes.
In \cite{xu}, smart sensors are assumed to be the recording devices and 
privacy preservation is considered to be the outcome of the proposed approach.
A Pareto evolutionary algorithm is proposed to optimize the average time consumption and average privacy entropy jointly.
In \cite{verma}, the authors consider the load balancing aspects of a model
that performs migration between Fog and Cloud infrastructures. The main techniques for 
achieving a fault tolerant system is replication and controlled migration. 
Finally, the authors of \cite{pus} elaborate on a clustering method for achieving load balancing in data migration. The presented scheme takes into consideration 
the network performance (e.g., bandwidth) targeting to  
minimize the transmission losses when a sudden traffic is observed.

The ultimate goal of the above presented efforts in the domain of synopses generation and services/data migration is to optimize the decision making upon a set of parameters.
Some common parameters are the time required to perform an action and the 
consumption of the available resources.
It is critical to deliver a scheme that efficiently drives the trade off between the cost of the action and the performance of the system.
Our current work differentiates from the discussed efforts in the sense that 
focuses on the continuous monitoring of the data similarity between processing nodes for future use in their decision making.
Actually, we enhance EC nodes with a reasoning mechanism that delivers 
the peers where every node can rely on when it needs their assistance in the 
envisioned processing activities.
The core idea is that data, due to the dynamic nature of the IoT environment and the mobility of users, could be updated and do not remain static. Hence, their statistics differ 
at consequent time instances. We do not propose neither another synopses generation method nor
a synopses/data migration scheme.
Our vision is to proactively know the (sub-)group of peers
that are the hosts of similar data and collaborate with them 
when data/services migration or tasks offloading is required.
The final outcome is a `map' of similar peers locally at every processing node combined with any 
synopses management technique or any services/data migration and tasks offloading 
mechanisms present in the relevant literature.

\section{Preliminaries \& High Level Description}
\label{preliminaries}
\begin{figure}[h!]
\centering
\includegraphics[width=0.4\textwidth]{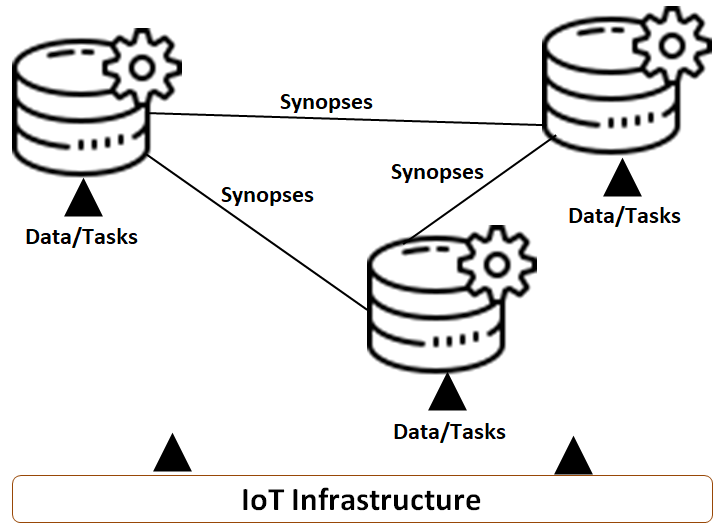}
\caption{Processing nodes sharing synopses while receiving data and tasks for execution.}
\label{fig:architecture}
\end{figure}

\textbf{Data \& Tasks Management at the Edge}.
The nomenclature adopted in this paper is presented by Table \ref{table:1}. 
Our setting involves $N$ processing nodes 
present at the EC, i.e., $\mathcal{N} = \left\lbrace n_{1}, n_{2}, \ldots, n_{N} \right\rbrace$ that have 
direct connection with 
a set of IoT devices that collect and report data. 
Upon these data, at every node $n_{i}$, a geo-distributed dataset is formulated 
that consists of 
a set of multivariate vectors, 
i.e., $\mathbf{x}^{i} = \left\lbrace x^{i}_{1}, x^{i}_{2}, \ldots, x^{i}_{M}\right\rbrace$ 
($M$ is the number of dimensions).
The discussed vectors depict the surrounding 
contextual information as recorded 
by IoT devices. 
EC nodes have the ability of communicating with their peers through the network
as presented by Fig. \ref{fig:architecture}. 
Apart from data, $n_{i}$ receives requests for various processing activities 
in the form of tasks 
relying on any arbitrary methodology (e.g., regression, classification, clustering 
and micro-clustering analysis, 
statistical dependencies between dimensions and so on and so forth). 
The aim is to generate knowledge and assist in the smooth execution of applications or 
fulfil users demands.
The collaborative nature of the aforementioned scenario imposes the exchange of information 
between nodes in order to have a view on the data present into their peers.
This is significant when a decision related to services/data migration or tasks offloading is the case.
However, it is not efficient to distribute humongous volumes of data in the network 
as this may jeopardise its performance and create bottlenecks.
In this paper, we propose the exchange of synopses instead of the `real' data to meet the discussed challenges.
A synopsis $\mathcal{S}^{i}$ is a summarization of 
a dataset present in the processing node $n_{i}$ depicting its statistical characteristics. 

$\mathcal{S}^{i}$ can be represented by a vector 
$\mathbf{s}^{i} = \left\langle s^{i}_{1}, s^{i}_{2}, \ldots, s^{i}_{l} \right\rangle \subset \mathbb{R}^{l}$
where $l$ is the number of dimensions of this vector.
We have to notice that $l$ can be equal to $M$ when 
$\mathbf{s}^{i}$ represents synopses for each dimension.
Some example synopses are as follows: \textbf{(i)} $\mathbf{s}^{i}$ can be the vector of coefficients in a linear multivariate regression model; 
\textbf{(ii)} If we focus on a clustering approach, $\mathcal{s}^{i}$ can depict the extracted centroids;
\textbf{(iii)} $\mathbf{s}^{i}$ can be the outcome of a simple statistical processing upon a sample of the local dataset like the generation of the mean or the standard deviation. 

An application scenario (where the adoption of synopses could play a significant role) may involve 
a group of unmanned 
vehicles moving around an area, collecting ambient contextual data
and performing simple processing activities.
A typical case is when unmanned vehicles 
execute some tasks for the monitoring of the environment.
In this scenario, it is necessary for unmanned vehicles to
distribute their data synopses to their peers in order to give them insights 
about the local data. When a vehicle is instructed to 
deliver, e.g., a machine learning model for a sub-area under surveillance, it 
may need data from its close peers to conclude a more generic model or
it may need to offload the task to another node that have a better view of 
the data/area.
However, the offloading should be at peers with which it exhibits 
a low distance (a high similarity) in the collected data otherwise the requestor
may receive wrong results. 
The low distance in data may 
naturally come in unmanned vehicles moving in the same sub-area while utilizing the same sensors
or may be exhibited due to the contextual environment (multiple nodes may detect the same phenomenon).

Nodes, at regular intervals, exchange the calculated synopses with their peers.
Our approach targets to build knowledge upon the similarity of nodes concerning their data 
in a temporal manner, i.e., we want to reveal the `stability' in the similarity 
between nodes upon the most recent observations. 
This way, our goal is to have a collaboration between nodes that, for a long time, record similar data as reported by IoT devices.
Additionally, we can easily detect the trade off upon the synopses exchange rate and the efficiency of the decision making.
The adoption of a low frequency will have  
consequences in the decision making due to the potential
inconsistencies between the real datasets and the reported synopses. 
The adoption of a high frequency will lead to an increased number of messages in the network negatively affecting the communication overhead.
A sophisticated decision making is then deemed necessary to optimally balance this trade-off taking into consideration the dynamic nature of the network and data contexts. 
This model is left for future work.   

\begin{table}[h!]
\centering
\label{nomenclature}
\caption{Nomenclature}
\begin{tabular}{ll}
\hline\hline
Notation & Description \\
\hline\hline
   $\mathcal{N}$  &  Set of processing nodes  \\
   $n_{i}$      &    the $i$th processing node         \\
   $\mathbf{x}$      &  $M$-dimensional data vector in $\mathbb{R}^{M}$               \\
   $\mathcal{S}^{i}$      & The synopsis produced in $n_{i}$                \\
   $\mathbf{s}^{i}$      &  $l$-dimensional synopsis vector in $\mathbb{R}^{l}$                \\
   $\alpha$ &  Least number of points in synopsis dominant clusters  \\
   $t$     &    Time instance for synopsis management     \\
	$\beta$ & Similarity value \\
	$d_{t}$ & The discrepancy quantum at $t$ \\
	$c_{k}$ & The centroid of the $k$th cluster \\
	$r_{k}$ & The radius of the $k$th cluster \\
	$m_{k}$ & The cardinality of the $k$th cluster \\
	$\overline{\beta}$ & The aggregated discrepancies as depicted by clusters \\
	$\beta$ & Similarity value \\
	$\xi$, $\zeta$, $\hat{\zeta}$ & Smoothing parameters \\
\hline\hline       
\end{tabular}
\label{table:1}
\end{table}

\textbf{High Level Description of the Proposed Model}.
The aim of this paper is to create a `temporal map' of the similarity between geo-distributed datasets as exposed by the processing of the most recent synopses. 
$n_{i}$ maintains a local data structure where the synopses of its peers are hosted.
The local decision making involves the processing of the received synopses adopting an unsupervised machine learning model (i.e., clustering) combined with a set of heuristics and 
a statistical measure (i.e., trend analysis).
Evidently, we will be able to have a view on the evolution of the similarity 
during time and depict the updates on the local datasets.
The temporal map can be adopted to deliver the list of peers that exhibit a high similarity 
with the local dataset and support decisions related to tasks and services management.
Our model keeps track of the evolution of data into peers and concludes 
an intelligent, data aware scheme for supporting the autonomous nature of EC nodes.

For such purposes, we define the synopses discrepancy quantum as the difference 
between two synopses at discrete time intervals.
Consider that $n_{i}$ receives a new synopses from $n_{j}$, i.e., $\mathbf{s}^{j}$. Naturally, $n_{i}$ relies on the $t \in \left\lbrace 1, 2, \ldots, W\right\rbrace$ latest reports to pay attention only on the recent updates of data into its peers.
The discrepancy quantum $d^{ij}_{t}$ is the discrepancy between 
synopses reported by the $i$th and $j$th nodes at the time instance $t$, i.e.,
$\mathbf{s}^{i}_{t}$ and $\mathbf{s}^{j}_{t}$.
For simplicity in our notation, from this point forward, we omit the indexes of
in the annotation of $\mathbf{s}_{t}$.
Based on the historical synopses reports, we take into consideration 
a time series $\left\lbrace d_{1}, d_{2}, \ldots, d_{W} \right\rbrace$ which depicts 
the track and trend of the discrepancy quantum.
The discrepancy quantum can be easily calculated by 
the L1 norm, i.e., 
$d_{t} = \sum_{k=1}^{M} |s^{i}_{k} - s^{j}_{k}|$ or
we can rely on other norms like the L2 norm, and so on and so forth.
We apply a clustering technique upon the time series of discrepancy quanta 
and deliver the indication of the temporal similarity between two nodes (it is performed for every pair of nodes) combined with the trend of the quanta upon the most recent quanta realizations. 
Our narrative in the upcoming sections refer in an individual node, however, the same rationale stands for every peer in the group.

\section{Temporal Management of Synopses}
\label{monitoring}

\textbf{Synopses Preparation}. 
Synopses are generated upon the collected data usually when new updates are received.
The interested reader can refer in \cite{aggarwal} for a survey 
of various 
methodologies (e.g., sampling, histograms, wavelets, sketches, micro-clusters).
Our model is not bounded or affected by the algorithm adopted to deliver the synopsis.
In our work, we adopt the synopses definition model presented in
\cite{kolomvatsostkde}, i.e., an online micro-clustering mechanism 
\cite{amini}.
This 
technique is built upon 
a tree-based structure focusing on the management of the \textit{Cluster Feature} $CF=\left\langle L, LS, SS \right\rangle$ concept. $CF$ is maintained for each 
cluster and contains the following information:
$L$ is the number of data points; 
$LS$ is their linear sum; 
and $SS$ is the square sum of data points. 
$CF$s are updated as new data arrive locally and the entire 
hierarchy is altered concerning the statistics of the affected part of the tree.
The generation of synopses 
refer in the extraction of statistics of 
internal tree-node(s) 
that represent clusters with at least $\alpha$ data vectors.
When the synopses update epoch expires, the node 
scans the $CF$-tree and retrieve 
the $\alpha$-dominant clusters (clusters with at least $\alpha$ data points). 
Let $I = median_{k}(L_{k})$ be the median and the median absolute deviation (MAD) of the number of points $L_{k}$ per cluster be $MAD = median_{k}(|L_{k}-I|)$.
We rely on clusters with at least $\alpha = I-3 \cdot MAD$ data points \cite{miller}. 
The presence of the remaining clusters is excluded 
from the synopses definition process. 

\textbf{Discrepancy Quanta Monitoring}. 
At $t$, $n_{i}$ receives $N-1$ synopses and updates 
the discrepancy quanta depicted by the time series 
$d_{1}, d_{2}, \ldots, d_{W}$. 
We rely on a sliding window approach where only the latest quanta are taken into consideration and
consider the discussed time series as the basis 
of a clustering mechanism that derives 
$C$ clusters. Every cluster depicts the grouping of the discrepancy quanta
being characterized by the vector $[c_{k}, r_{k}, m_{k}]$ where
$c_{k}$ is the corresponding centroid, $r_{k}$ is the radius (the distance between the centroid and the 
most distant object in the cluster) and 
$m_{k}$ is the number of the discrepancy quanta at the $k$th cluster
(i.e., the number of objects in the cluster - the cardinality of the cluster).
The clustering process
represents the distribution of the discrepancy quanta in the time series
and gives an indication about the evolution of the similarity 
between datasets present in different nodes.
Obviously, when $d \to 0$ indicates a high similarity and the opposite stands 
when $d \to \infty$.
Upon the $C$ clusters, we have to deliver the final similarity for the current 
window $W$.
Every centroid corresponds to a different discrepancy quantum representation, thus, to a different similarity value.
To cover this uncertainty, we consider an heuristic to deliver the final similarity value $\beta$.
The ideal case is to detect only one cluster containing all the discrepancy quanta 
with a strong correlation upon $c_{k}$. The strong correlation can be represented by 
the placement of discrepancy quanta in a limited radius around $c_{k}$.
The shorter the radius is, the higher the `belief' on the $c_{k}$ becomes.
The higher the cardinality is, the higher the `belief' on the specific cluster becomes.
A cluster containing a high number of discrepancy quanta in a limited radius should affect more the final $\beta$ as it consists of a representative group of values concerning the similarity of the studied synopses, thus, the corresponding datasets.
 
We define the function $f(\cdot)$ to deliver the final value of the discrepancy,
i.e., $\overline{\beta} = f\left( \left\lbrace c_{k}, r_{k}, m_{k} \right\rbrace_{k=1}^{C} \right)$.
$f(\cdot)$ aggregates the `contribution' of each cluster to the final realization of 
$\beta$.
The following equation holds true:
\begin{equation}
f\left( \left\lbrace c_{k}, r_{k}, m_{k} \right\rbrace_{k=1}^{C} \right) = \sum_{k=1}^{C} c_{k} w_{k}
\end{equation}
where $w_{k} = \frac{\overline{w}_{k}}{\sum_{\forall k} \overline{w}_{k}}$ is the weight for each centroid.
We strategically decide to adopt a very simple heuristic to conclude $\overline{w}$,
i.e., $\overline{w}_{k} = \frac{m_{k}}{r_{k}}$. 
Based on $\overline{\beta}$, we can easily define the similarity between the two datasets as follows:
\begin{equation}
\beta = \frac{1}{1+e^{\xi \overline{\beta} - \zeta}}
\end{equation} 
where $\xi, \zeta \in \mathbb{R}^{+}$ are smoothing parameters.
$\beta$ is defined in the unity interval and approximates zero when the final discrepancy as exposed by the aforementioned clusters is over a threshold. Naturally, a high discrepancy will lead to a low similarity, thus, the corresponding peer does not own a dataset that can be used in case where data migration or tasks offloading should be decided.

\textbf{Discrepancy quanta trend management}.
We propose the use of a non parametric statistical trend analysis upon 
the discrepancy quanta $d_{t}$ and combine it with the 
similarity extraction mechanism based on machine learning presented in the previous paragraph.
Through the proposed approach, we try to combine two different models 
to reveal if there is a high similarity between two datasets and it is maintained in 
the recent observations. 
Our trend analysis is performed only for the recent $\eta \cdot W$ time instances and not for the entire
historical window.
The recent variability of discrepancy quanta will be smoothly aggregated (discussed later) with the 
outcome of the machine learning model. 
Our trend analysis is similar to the model presented in one of our past efforts
\cite{kolomvatsosicics} and is based on 
the widely known Mann-Kendall metric or Mann-Kendall test (MKM) \cite{kendall}, \cite{mann}.
The MKM
is adopted to 
indicate if there are trends in a time series sequence, i.e., $\left\lbrace d_{i} \right\rbrace, i=1,2, \ldots, \eta \cdot W$, $\eta \in (0,1]$.
MKM pays attention on the sign of 
the difference between the observed data and previous measurements and compares every later-measured data with previous observations in pairs.
The significant is that the MKM is invariant to transformations (e.g., logs) enhancing its applicability in multiple application domains.
The following equation holds true:
\begin{equation}
 S = \sum_{i=(1-\eta)W+1}^{W - 1} \sum_{j=i+1}^{W} sign(d_{j} - d_{i})
\end{equation}
where $sign(d_{j} - d_{i})$ is considered equal to unity if $d_{j} > d_{i}$, equal to zero if $d_{j} = d_{i}$ and equal to -1 when $d_{j} < d_{i}$. Upon $S$, we can define the parameter $Z$ realized as follows:
$Z = \frac{S-1}{var(S)}$ if $S>0$, $Z = 0$ if $S=0$ and $Z = \frac{S+1}{var(S)}$ if $S<0$
where $var(S) = \frac{S\cdot(q-1)\cdot(2q+5)- \sum_{i=1}^{q'} tp\cdot(tp-1)\cdot(2tp+5)}{18}$,
$q = \eta \cdot W$, $tp$ is the ties of the $p$th value and $q'$ is the number of ties.
Finally, if $Z$ is positive, we can conclude an increasing trend and the opposite stands when $Z$ is negative. 

\textbf{Trend Aware Similarity Realization}.
As noted, a positive realization of the variable $Z$ depicts and increasing trend in the discrepancy quanta while the opposite stands for a negative $Z$. Obviously, a negative $Z$ in combination with 
a high $\beta$ leads to the conclusion of a high similarity between the two nodes.
For aggregating 
the outcome of the machine learning with the result of the trend analysis, we rely on a simple, however, fast heuristic. The following equation stands true: $\hat{\beta} = \epsilon \cdot \beta$ where
$\epsilon = \frac{1}{e^{\hat{\zeta} \cdot Z}}$ if $Z<0$, $\epsilon = e^{-\hat{\zeta}\cdot Z}$ if $Z>0$ and $\epsilon = 1$ if $Z=0$ with $\hat{\zeta}$ being a smoothing parameter.
$\epsilon$ is adopted to increase/decrease $\beta$ based on the most recent trend of the discrepancy quanta. Obviously, the proposed model is applied for every peer node and, finally, produces a sorted list based on 
$\hat{\beta}$. 
Hence, when the current node decides, e.g., to offload a task or `borrow' data from its peers, it can rely on the aforementioned list and interact with peers that exhibit the highest similarity value (e.g., a top-$k$ model).

\section{Performance Evaluation}
\label{evaluation}
The proposed model, i.e., the Synopses Discrepancy Management Model (SDMM), is evaluated through the adoption of a wide set of experimental scenarios upon a number of performance metrics. In the following paragraphs, we provide details about the experimental setup, the considered metrics and the 
performance outcomes.

\textbf{Performance Metrics}.
We, initially, study the discrepancy quantum at $t=W$ where the decision for the final similarity between datasets
is made. We define the $\gamma$ metric that depicts the discrepancy quantum at $t=W$, i.e.,
$\tilde{d}_{W}$, compared to the minimum discrepancy quantum in the 
group of peers $\hat{d}_{W}$. The following equation stands true:
$\gamma = \frac{\tilde{d}_{W}}{\hat{d}_{W}}$.
When $\gamma \to 1$ means that the proposed model manages to select the 
best possible peer to be in the first place of the sorted list being directly adopted when 
services/data migration or tasks offloading is necessary.
When $\gamma >> 1$ means that the proposed model does not conclude the peer with the lowest discrepancy at $t$ a the first place of the final sorted list.
The selection of the minimum discrepancy quantum at $W$ consists of the baseline model 
for our method as it performs a greedy approach based only on the latest view of the similarity 
without paying any attention on historical values and the trend of the discrepancy. However, the baseline 
model does not perform well when the trend of the discrepancy quantum is positive being increased at 
the next observation made at $t=W+1$. 
We also define the metric $\delta$ that represents the `top-$k$' group of peers as exposed by our model 
compared to the `minimum-$k$' discrepancy quanta at $t=W$. 
$\delta$ is calculated as follows:
$\delta = \frac{|K \cap K_{baseline}|}{k}$ where $K$ is the set of the top-$k$ nodes present into the final sorted list and $K_{baseline}$ is the $k$ peers with the minimum discrepancy at $t=W$.
For depicting the effect of the trend of the discrepancy quanta into the envisioned decision making, we define 
$\omega$ \& $\epsilon$ metrics.
$\omega$ represents the ability of the model to detect and select peers with a `negative' trend while $\epsilon$ reports on 
the selection of peers that exhibit a `positive' trend in the discrepancy quanta.
We observe the event of a peer being in the top-$k$ similarity list, i.e.,
$\tilde{E}$ and the event of a negative discrepancy quantum, i.e.,
$\hat{E}$. The following equations hold true:
$\gamma = |\tilde{E} \& \hat{E}|$, $\delta = |\tilde{E} \& \neg \hat{E}|$ where 
$||$ depicts the number (cardinality) of the observations for the specific conjunctions of the aforementioned events.
Obviously, $\gamma$ \& $\delta$ are complement each other and the best performance is exhibited by a high value of $\gamma$ (low value of $\delta$).  

\textbf{Datasets \& Parameters}. 
We experiment with two real datasets: 
\textbf{(i)} Dataset D1 \cite{Aliaj} contains 
three-dimensional ($M=3$) vectors with TCP network performance metrics\footnote{https://mm.aueb.gr/unsurpassed/}, i.e., throughout (MBit/s), size of TCP congestion window (Kbytes), and link cost, recorded every 10s by a swarm of ten Unmanned Surface Vehicles (USVs). Each USV, equipped with a Raspberry Pi for local computation, floats on sea surface in the coastal area of Skaramagas, Greece, communicating using TCP. 
The pdf of throughput and congestion metrics with Coefficient of Variation (CoV) 0.80 and 1.15, resp., indicating highly variance variables (CoV is std deviation to mean ratio $\frac{\sigma}{\mu}$). Congestion is strongly, positively correlated with throughput (Pearson coefficient 0.79), while the link cost has negative moderate correlation with congestion and throughput (-0.56 and -0.43, resp.);   
\textbf{(ii)} Dataset D2 \cite{natascha} contains two-dimensional USVs sensor readings \footnote{https://sites.google.com/view/gnfuv}, i.e., 
humidity \& temperature
recorded every 10s by a swarm of four USVs
floating in the coastal area of Skaramagas, Greece. 
The CoV for temperature and humidity is 0.11 and 0.09, resp. with Pearson coefficient -0.41, indicating negative moderate correlated variables with relatively low variance. 
Upon those datasets, we generate 
the required traces adopted in our simulations.
We also provide a comparative assessment of our SDMM with 
a Moving Average (MA) model\footnote{https://bookdown.org/JakeEsprabens/431-Time-Series/}. The MA 
is a time series model that accounts for very short run autocorrelation. The MA assumes that the next observation is the mean of a number of historical values. In our experiments, we consider the last four observations of discrepancy quanta as the 
time series for retrieving the outcome of the MA.
After every experiment, 
we take the mean values of the above described metrics over 1,000 runs, $W \in \left\lbrace 10, 100, 1000 \right\rbrace$,
$N=10$,
$\eta \in \left\lbrace 0.5, 1.0 \right\rbrace$, $\xi = 2.0$, $\zeta = \hat{zeta} = 35.0$. 
At each run, we construct the necessary synopses, calculate the discrepancy quanta and apply our mechanism. 

\begin{figure}[h!]
\centering
\includegraphics[width=0.55\textwidth]{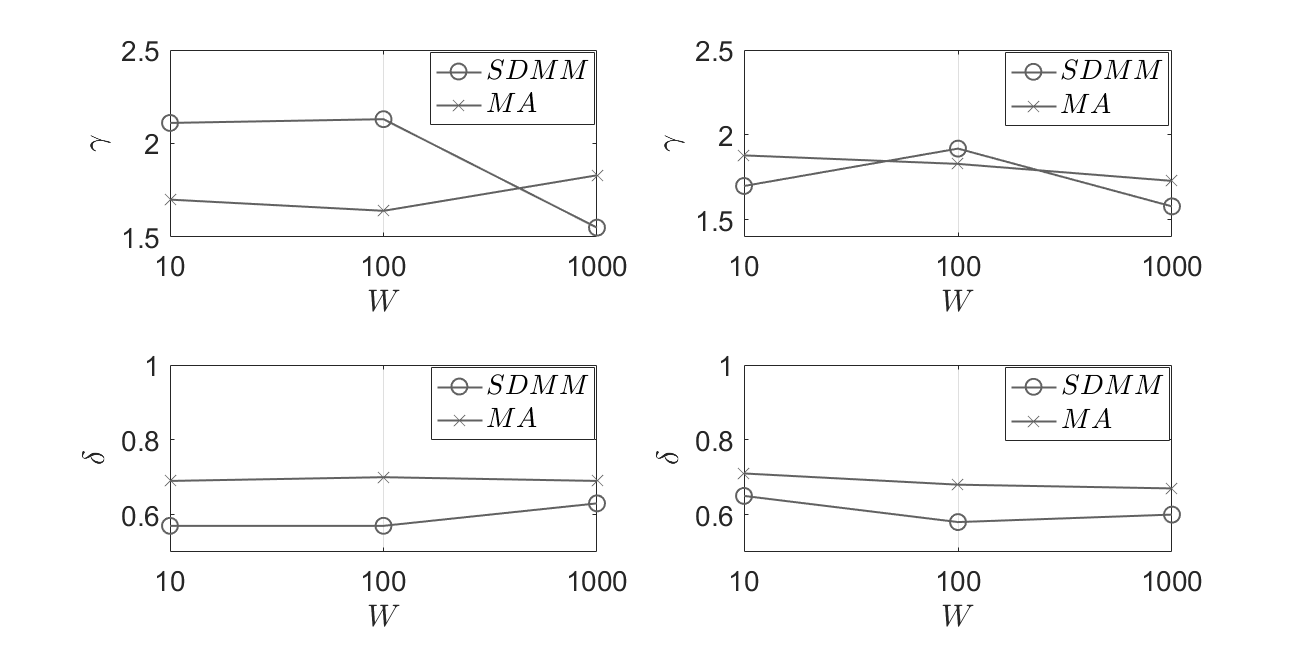}
\caption{Performance outcomes for $\gamma$ \& $\delta$ (dataset: D1 - left: $\eta=0.5$, right: $\eta=1.0$).}
\label{fig1}
\end{figure}

\textbf{Performance Assessment}.
Initially, we experiment with  
D1 and depict the outcomes for the above described metrics.
In Fig. \ref{fig1}, we present the evaluation results for $\gamma$ \& $\delta$ metrics. We observe that the proposed model outperforms the MA when the size of the sliding window is high (i.e., $W \to 1000$).
Then, the SDMM is capable of detecting and locating at the first place of the final sorted list a peer that exhibits a discrepancy quantum close to the minimum possible.
The parameter $\eta$ also affects the performance. When $\eta = 0.5$, i.e., we consider a part of the entire window (the half of the window) for trend analysis, the SDMM exhibits a slightly worse performance especially for low $W$ ($W \in \left\lbrace 10, 100 \right\rbrace$) than the MA.
This means that the proposed approach is not fed with the 
appropriate amount of data to discern the hidden trend behind the discrepancy quanta realizations.
The specific sun-window taken into consideration does not 
fully depict the sufficient statistics to discern the trend of the time series.
Concerning $\delta$, we observe a similar performance no matter the $\eta$ value. In general, 60\% of the detected peers are the nodes exhibiting the minimum discrepancy quanta at $W$. 
The SDMM exhibits worse performance than the MA when we focus on $\delta$. Naturally, the MA relying only on the latest four quanta realizations manages to detect more 
peers located in the `minimum-$k$' list. 
The difference in the performance between the SDMM and the MA can be observed in the interval [10\%, 15\%] in favour of the MA which does not represent a scenario where the SDMM cannot be adopted in a real setup. We reach to this conclusion if we consider that the SDMM does not support the decision making only on the 
latest observations but in the entire list of historical measurements to feed the unsupervised 
learning model, i.e., we try to capture future insights on discrepancies. 
This approach better matches to our needs for detecting 
the evolution of discrepancy quanta through time. Recall that the outcome of the machine learning scheme is combined with the statistical trend analysis for delivering a holistic model to sufficiently identify 
the statistics of quanta.

\begin{figure}[h!]
\centering
\includegraphics[width=0.55\textwidth]{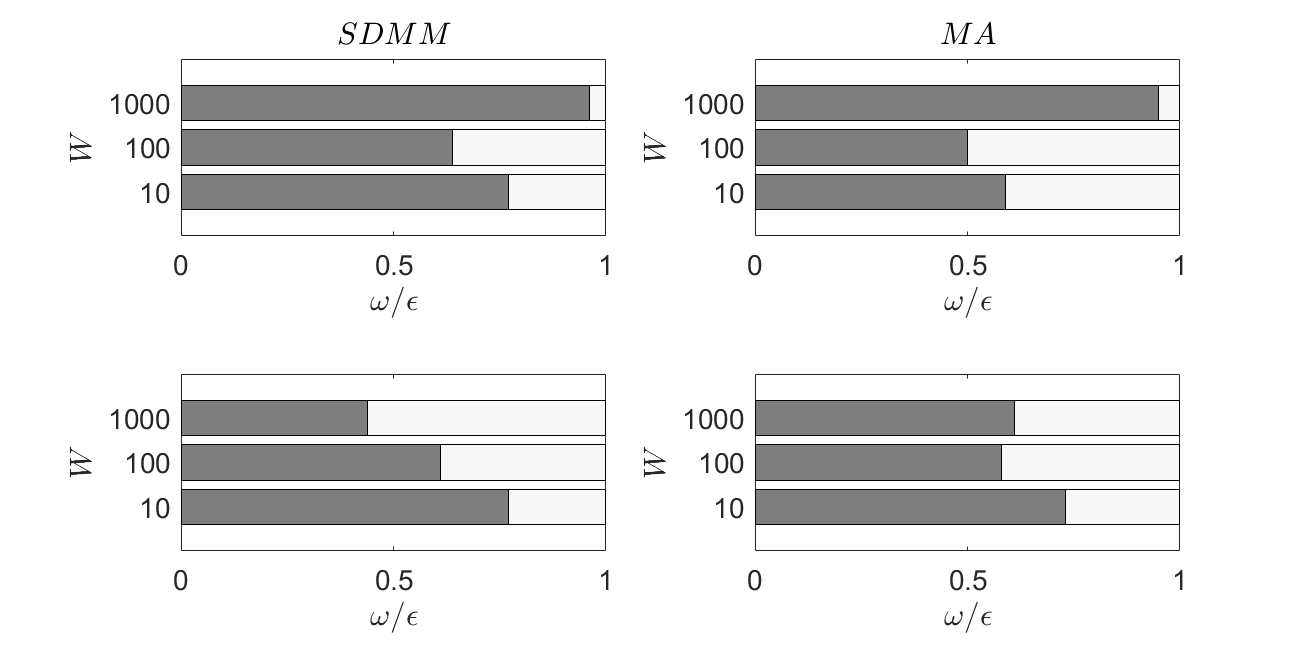}
\caption{Performance outcomes for $\omega$ \& $\epsilon$ - D1 (left: SDMM, right: MA - 1st row: $\eta=0.5$, 2nd row: $\eta=1.0$).}
\label{fig2}
\end{figure}

In Fig. \ref{fig2}, we provide our outcomes for 
$\omega$ \& $\epsilon$ metrics.
At the left, we depict the performance of the SDMM and, at the right, we present the performance of the MA.
The first row of the figure concerns the scenario where $\eta = 0.5$ and the second row the scenario where 
$\eta=1.0$. The SDMM outperforms the MA except the scenario where $\eta=1.0$ \& $W=1000$.
In general, the proposed scheme manages to detect and select peers that exhibit a `negative' trend in the observed discrepancy quanta in combination with the lowest possible discrepancies. 
Our model depicts the knowledge upon the evolution of discrepancies that will be adopted 
for decision making at the future.
The discussed results depict our intention to take into consideration not only a low discrepancy quantum but also the trend in the evolution of quanta through time. 
A large window $W$ gives the opportunity to better detect the statistical characteristics of the discrepancies 
resulting the highest possible $\omega$ when $\eta=0.5$.
However, a very high window may `disturb' the trend analysis as depicted by the experimental scenario where
$\eta=1.0$ \& $W=1000$ especially if we consider dynamic environments like those represent by the adopted traces. 
Such disturbances are observed in dynamic environments being depicted by changes in the underlying data.

\begin{figure}[h!]
\centering
\includegraphics[width=0.55\textwidth]{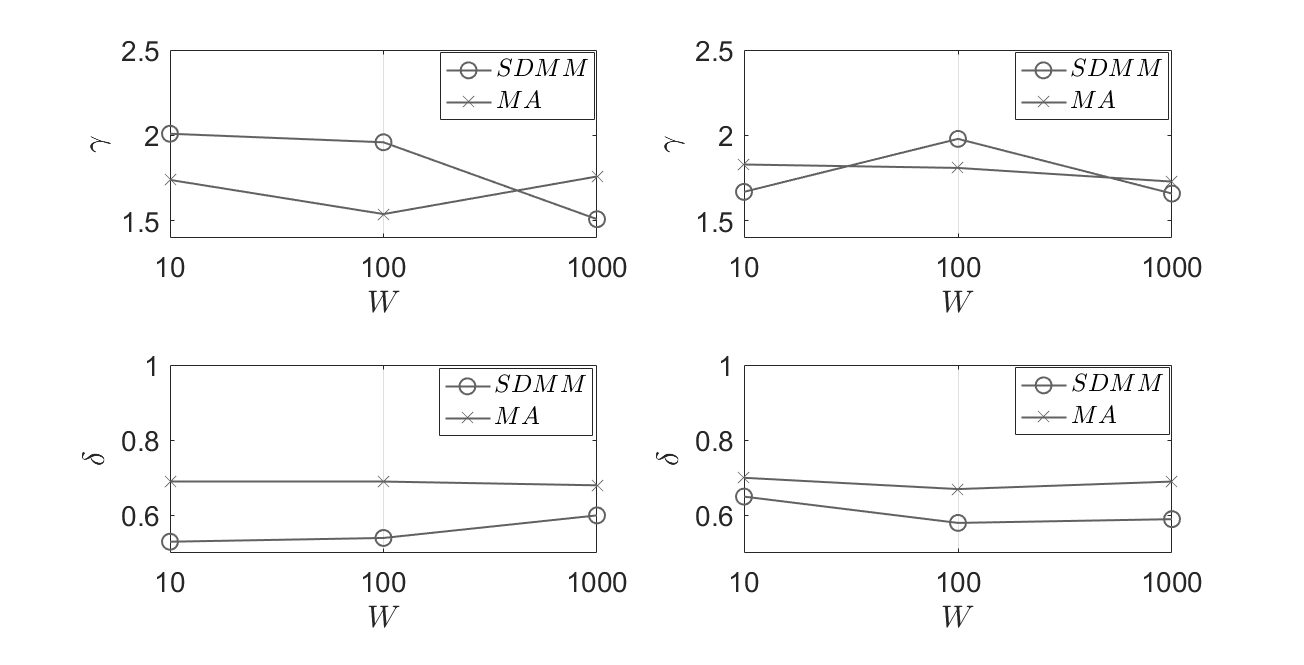}
\caption{Performance outcomes for $\gamma$ \& $\delta$ (dataset: D2 - left: $\eta=0.5$, right: $\eta=1.0$).}
\label{fig3}
\end{figure}

\begin{figure}[h!]
\centering
\includegraphics[width=0.55\textwidth]{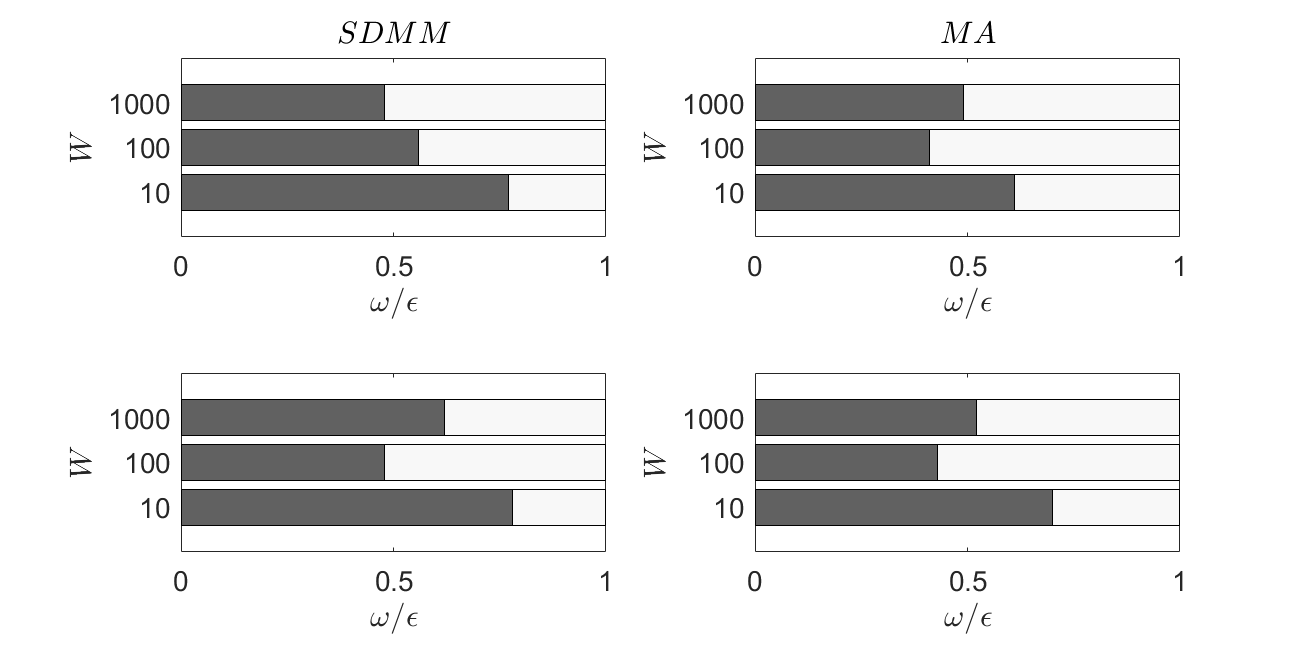}
\caption{Performance outcomes for $\omega$ \& $\epsilon$ - D2 (left: SDMM, right: MA - 1st row: $\eta=0.5$, 2nd row: $\eta=1.0$).}
\label{fig4}
\end{figure}

We proceed with the experimentation upon D2 and present our results in Figs \ref{fig3} \& \ref{fig4}.
Again, we observe similar outcomes like in the scenario where D1 feeds our models.
The adoption of a high $\eta$ leads to better performance
for a low $W$ concerning $\omega$/$\epsilon$, however, the performance is worst for an increased window size. The reason behind that is the statistics of the adopted dataset and the updates on the collected data. Now, the SDMM exhibits worse performance than the MA in a different scenario (compared to the experimentation performed for D1), i.e., when
$\eta=0.5$ \& $W=1000$. The mean difference, in favour of the SDMM, concerning $\omega$ is 15\% (approx.) and 12\% (approx.) for $\eta=0.5$ \& $\eta=1.0$, respectively.

\section{Conclusions \& Future Work}
\label{conclusions}
Data management at the edge of the network opens up the room for 
building intelligent applications upon the data collected by numerous devices present at the Internet of Things infrastructure.
The ability of edge nodes to collaborate in data processing or the execution of various tasks exposes the need for efficient mechanisms that will facilitate the communication and collaboration between edge nodes. 
In fact, we need models that incorporate the knowledge about the status of peers, thus, any edge node is capable of interacting with the appropriate peers when necessary.
In this paper, we elaborate on a scheme that prepares a temporal similarity map between the datasets present at the edge ecosystem. We want to support the decision making at every node locally and expose the peers where similar data are present.
We build upon the concept of the discrepancy quantum and provide the relevant monitoring mechanism for it.
The target is to collect the historical discrepancy quanta and depict the estimated long term similarity and its trend between datasets owned by different edge nodes. 
We present the formulation of the problem and the details of our solution.
The evaluation of our model is realized through the adoption of real datasets and a set of performance metrics. 
We show that our approach can be easily adopted to perform the desired goals and is capable of revealing the discrepancy in the exchanged data synopses and deliver 
the envisioned similarity map.
In the first place of our future research plans is the adoption of 
a deep learning model for detecting the hidden statistical characteristics of discrepancy quanta before we conclude the final decision about the similarity level between two different datasets.


\begin{thebibliography}{00}


\bibitem{Aggarwal1}
Aggarwal, C., et al., 'A Framework for Clustering Evolving Data Streams', VLDB, 2003.


\bibitem{aggarwal}
Aggarwal, C., Yu, P., 'A Survey of Synopsis Construction in Data Streams', ch. in 'Data Streams, Models and Algorithms', Springer Science \& Business Media, 2007. 

\bibitem{alam}
Alam, M., et al., 'Multi-agent and reinforcement learning based code offloading in mobile fog', in ICOIN, 2016, 285-–290.

\bibitem{alfarraj}
Alfarraj, O., 'A machine learning-assisted data aggregation and offloading systemfor cloud–IoT communication', Peer-to-Peer Netw. Appl., 2020. 

\bibitem{Aliaj}
Aliaj, E., et al. 'A platform for wireless maritime networking experimentation', GIIS, 2018.

\bibitem{amini}
Amini, A., Wah, T. Y., 'Density Micro-Clustering Algorithms on Data Streams: A Review', Intl. Conf. Engineers \& Computer Scientists, 2011.


\bibitem{anagnotson}
Anagnostopoulos, C., Hadjiefthymiades, S., Kolomvatsos, K., 'Accurate, Dynamic \& Distributed Localization of Phenomena for Mobile Sensor Networks', ACM TOSN, 12(2), No 9, 2016.

\bibitem{anagnogrouping}
Anagnostopoulos, C., Hadjiefthymiades, S., Kolomvatsos, K., 'Time-optimized user grouping in Location Based Services', Computer Networks, 81, 2015, 220--244. 

\bibitem{Babcock}
Babcock, B., et al., 'Models and issues in data stream systems', PODS, 2002.

\bibitem{bellavista}
Bellavista, P., et al., 'Differentiated Service/Data Migration for Edge Services Leveraging Container Characteristics', IEEE Access, 7, 2019.

\bibitem{bellavista2}
Bellavista, P., Zanni, A., Solimando, M., 'A migration-enhanced edge computing support for mobile devices in hostile environments', in 13th IWCMC, 2017, 957–-962.


\bibitem{Breitbach}
Breitbach, M., et al., 'Context-Aware Data and Task Placement in Edge Computing Environments', IEEE PerCom, 2019.

\bibitem{carrega}
Carrega, A., et al., 'A Middleware for Mobile Edge Computing', IEEE Cloud Computing, 4(4), 2017, 26—37.


\bibitem{Chakrabarti}
Chakrabarti K., et al., 'Approximate Query Processing with Wavelets', VLDB J, 10(2-3):199--223, 2001.

\bibitem{chen2}
Chen, X., et al., 'Optimized computation offloading performance in virtual edge computing systems via deep reinforcement learning', IEEE Internet of Things, vol. 6(3), pp. 4005--4018, 2019.



\bibitem{devita}
De Vita, F., et al., 'A Deep Reinforcement Learning Approach for Data Migration in Multi-Access Edge Computing', Machine Learning for a 5G Future (ITU K), 2018, 10.23919/ITU-WT.2018.8597889.

\bibitem{dinh}
Dinh, T., et al., 'Offloading in mobile edge computing: Task allocation and computational frequency scaling', IEEE Transactions on Communications, vol. 65, no. 8, pp.3571--3584, 2017.


\bibitem{natascha}
Harth, N., Anagnostopoulos, C. 'Quality-aware aggregation \& predictive analytics at the edge', IEEE Big Data 2017, 17--26.

\bibitem{hong}
Hong, L., 'The Absolute Difference Law For Expectations', The American Statistician, 69:1, 8-10, 2015, 10.1080/00031305.2014.994712.

\bibitem{huang}
Huang, L., et al., 'Deep reinforcement learning for online computation offloading in wireless powered mobile-edge computing networks', IEEE Transactions on Mobile Computing, vol. 19(11), 2019, pp. 2581--2593.

\bibitem{karanika}
Karanika, A., Oikonomou, P., Kolomvatsos, K., Loukopoulos, T., 'A Demand-driven, Proactive Tasks Management Model at the Edge', IEEE FUZZ-IEEE, 2020.

\bibitem{kendall}
Kendall, M., 'Rank Correlation Methods', Charles Griffin, 1975.

\bibitem{kolomvatsosCOMPUTING}
Kolomvatsos, K., 'A Distributed, Proactive Intelligent Scheme for Securing Quality in Large Scale Data Processing', Computing, 2019.

\bibitem{kolomvatsosFGCS}
Kolomvatsos, K., Anagnostopoulos, A., 'Multi-criteria Optimal Task Allocation at the Edge', Future Generation Computer Systems, 93:358--372, 2019.

\bibitem{kolomvatsosiot}
Kolomvatsos, K., et al., 'Distributed Localized Contextual Event Reasoning under Uncertainty', IEEE Internet of Things Journal, 4(1), 2017, 183-191.

\bibitem{kolomvatsosfusion}
Kolomvatsos, K., et al., 'Data fusion and type-2 fuzzy inference in contextual data stream monitoring', IEEE TSMC:Systems, 47(8), 2016, 1839--1853. 

\bibitem{kolomvatsostopk}
Kolomvatsos, K., et al., 'A time optimized scheme for top-k list maintenance over incomplete data streams', Information Sciences, 311, 2015, 59--73.

\bibitem{kolomvatsosiisa}
Kolomvatsos, K., et al., 'An efficient environmental monitoring system adopting data fusion, prediction, \& fuzzy logic', 6th IISA, 2015.

\bibitem{kolomvatsostkde}
Kolomvatsos, K., et al., 'Proactive \& Time-Optimized Data Synopsis Management at the Edge', IEEE TKDE, 2020.

\bibitem{kolomvatsosicics}
Kolomvatsos, K., Anagnostopoulos, C., 'Landmark based Outliers Detection in Pervasive Applications', in 12th International Conference on Information and Communication Systems, 2021.

\bibitem{kolomvatsosPSO}
Kolomvatsos, K., Hadjiefthymiades, S., 'On the Use of Particle Swarm Optimization and Kernel Density Estimator in Concurrent Negotiations', Information Sciences, 262, 2014, 99--116.

\bibitem{kolomvatsossemantic}
Kolomvatsos, K., et al., 'Semantic location based services for smart spaces', Metadata and Semantics, Springer, 2009, 515--525.

\bibitem{Lakshmi}
Lakshmi, K. P., Reddy, C. R. K., 'A Survey on Different Trends in Data Streams', IEEE Intl Conf Netw. \&  Inf. Techn., 2010. 

\bibitem{mann}
Mann, H., 'Nonparametric tests against trend', Econometrica 13: 245-259, 1945.


\bibitem{miller}
Miller, J. 'Reaction time analysis with outlier exclusion: Bias varies with sample size', J Exper. Psych. 43A(4):907--912, 1991.

\bibitem{najam}
Najam, S., et al., 'The Role of Edge Computing in Internet of Things', IEEE Communications Magazine, 2018.

\bibitem{pus}
Puschmann, D., et al., 'Adaptive clusteringfor dynamic IoT data streams', IEEE Internet Things, 1-1, 2016.



\bibitem{expectation}
Simon, M., 'Maximum and Minimum of Pairs of Random Variables', In: Probability Distributions Involving Gaussian Random Variables. A Handbook for Engineers and Scientists, Springer, Boston, MA, 2002. 

\bibitem{Tatbul}
Tatbul, N., Zdonik, S., 'A subset-based load shedding approach for aggregation queries over data streams', VLDB 2006.

\bibitem{verma}
Verma, S., et al., 'An efficient data replication and load balancing technique for fog computing environment', in 3rd ICCSGD, 2016, 2888--2895. 


\bibitem{wangsurvey}
Wang, S., et al., 'A Survey on Service Migration in Mobile Edge Computing', IEEE Access, PP(99), 2018.

\bibitem{xu}
Xu, Z., et al., 'A time-efficient data offloading method with privacy preservation for intelligent sensors in edge computing', EURASIP Journal on Wireless Communications and Networking, No 236, 2019.


\bibitem{zhang}
Zhang, T., et al., 'Data offloading in mobile edge computing: A coalitional game based pricing approach', IEEE Access, PP(99), 2017.


\end{thebibliography}
\end{document}